\documentclass[english]{article}
\usepackage[latin9]{inputenc}
\usepackage{geometry}
\usepackage{float}
\usepackage{amsmath}
\usepackage{graphicx}

\makeatletter

\providecommand{\tabularnewline}{\\}
\newcommand{\lyxdot}{.}

\floatstyle{ruled}
\newfloat{algorithm}{tbp}{loa}
\providecommand{\algorithmname}{Algorithm}
\floatname{algorithm}{\protect\algorithmname}

\date{September 2004}

\makeatother

\usepackage{babel}
\begin{document}

\title{Global optimization using L\'evy flights}

\author{Truyen Tran, Trung Thanh Nguyen, Hoang Linh Nguyen}
\maketitle
\begin{abstract}
This paper studies a class of enhanced diffusion processes in which
random walkers perform L\'evy flights and apply it for global optimization.
L\'evy flights offer controlled balance between exploitation and
exploration. We develop four optimization algorithms based on such
properties. We compare new algorithms with the well-known Simulated
Annealing on hard test functions and the results are very promising.\\
\\
\textbf{Key words}: L\'evy flights, global optimization.\\
\emph{This is an edited version of a paper originally published in
Proceedings of Second National Symposium on Research, Development
and Application of Information and Communication Technology} (ICT.rda'04\emph{)},
Hanoi, Sept 24-25, 2004.
\end{abstract}

\section{Introduction}

Optimization can be characterized as a process to minimize an objective
function over a bounded or unbounded space. Of the widely used class
of optimization algorithms known as meta-heuristics, that many of
them imitate natural processes such as Simulated Annealing (SA - originated
in physics), Genetic Algorithm (GA - imitating the Darwinian process
of natural selection) \cite{De-Jong75}, Ant Colony Optimization (ACO
\textendash{} mimicking behavior of foraging ants) \cite{dorigo1996ant}
and Particle Swarm Optimization (PSO \textendash{} modeling the flocking
and schooling by birds) \cite{eberhart2001particle}. In particular,
SA and GA have been practically successful although they never guarantee
to reach global optima within limited time.

This paper studies a group of stochastic processes frequently observed
in physics and biology called \emph{L\'evy flights} \cite{bardou2000cooling}\cite{viswanathan1999optimizing}.
The goal is to realize the claim by Gutowski that L\'evy flights
can be used in optimization \cite{gutowski2001evy}. A new class of
meta-heuristic algorithms called LFO (L\'evy Flights Optimization)
is introduced. To our knowledge, similar work has not been introduced
in the optimization literature.

The rest of paper is organized as the following. Sec.~2 outlines
some fundamentals of L\'evy flights. Sec.~3 introduces four new
LFO algorithms. Sec.~4 reports experiments and results. The last
section provides several possible outlooks.

\section{L\'evy flights \label{sec:Lvy-flights}}

Extensive investigations in diffusion processes have revealed that
there exist some processes not obeying Brownian motion. One class
is \emph{enhanced diffusion}, which has been shown to comply with
L\'evy flights \cite{viswanathan1999optimizing}. The L\'evy flights
can be characterized by following probability density function:
\begin{eqnarray}
P(x) & \sim & \left|x\right|^{-1-\beta}\,\,\mbox{as}\,\, x\rightarrow\infty,\,\,\mbox{where}\nonumber \\
0 & <\beta\le & 2\label{eq:Levy}
\end{eqnarray}
Note that with $\beta\le0$, the distribution in Eq.~(\ref{eq:Levy})
cannot be normalized, therefore it has no physical meaning although
our computation needs not be concerned about such problem. For $0<\beta<1$,
the expectation does not exist.

\begin{figure}

\centering{}%
\begin{tabular}{cc}
\includegraphics[width=0.5\textwidth,height=0.4\textwidth]{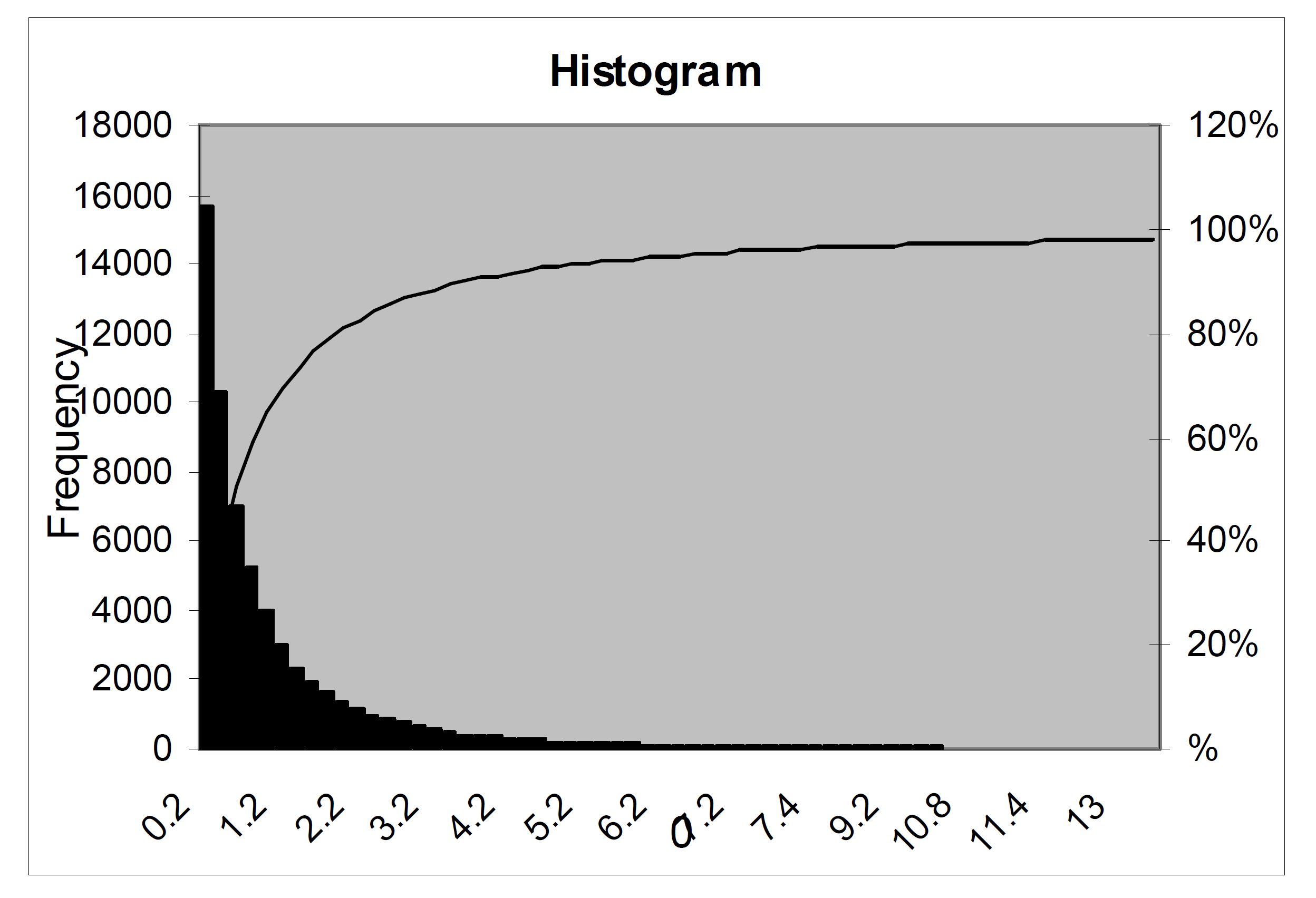} & \includegraphics[width=0.5\textwidth,height=0.4\textwidth]{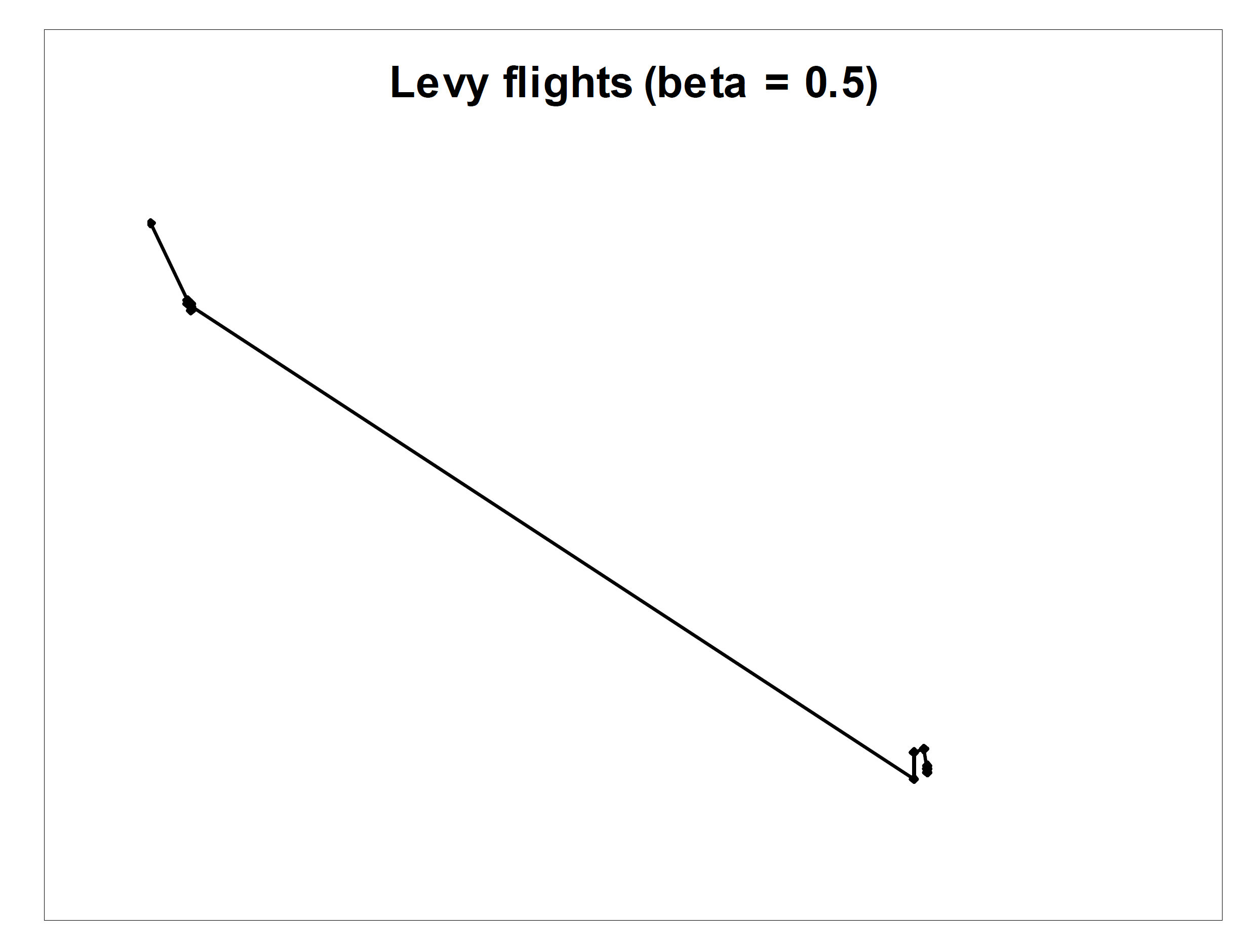}\tabularnewline
(a) $\beta=1.5$ & (b) $\beta=0.5$\tabularnewline
\includegraphics[width=0.5\textwidth,height=0.4\textwidth]{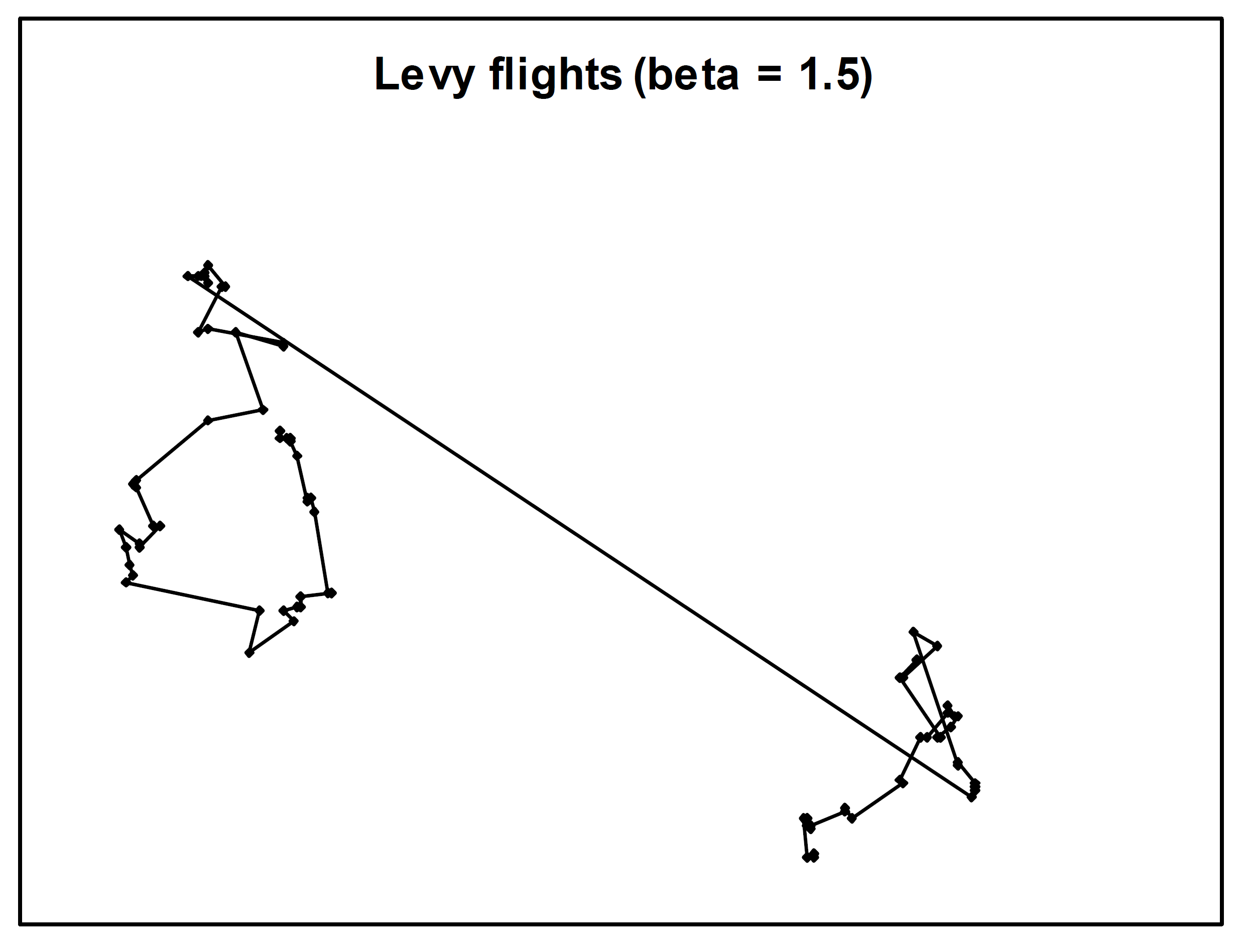} & \includegraphics[width=0.5\textwidth,height=0.4\textwidth]{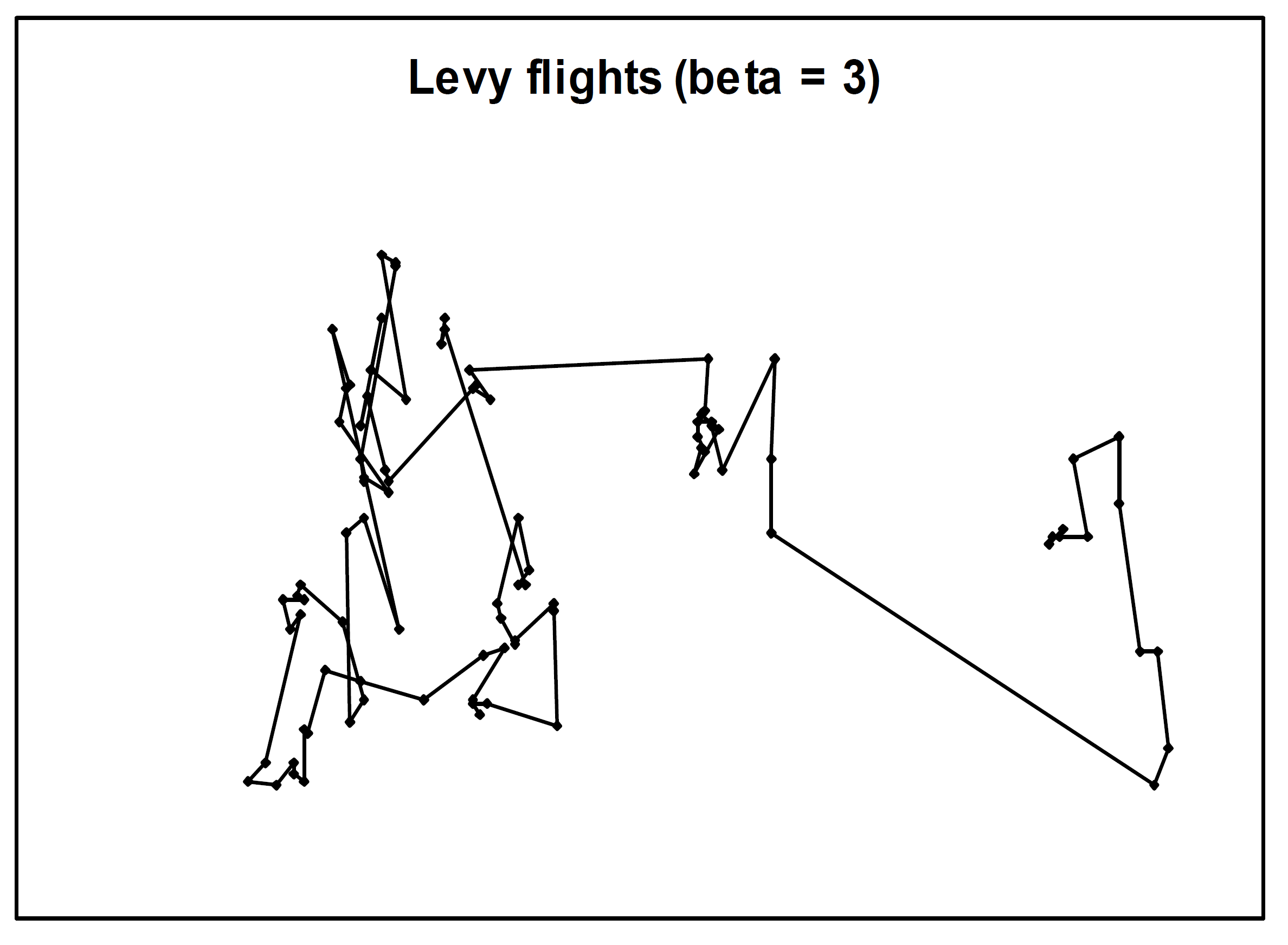}\tabularnewline
(c) $\beta=1.5$ & (d) $\beta=3.0$\tabularnewline
\end{tabular}\caption{Flight length distributions with various $\beta$. \label{fig:Flight-length-distribution}}
\end{figure}

For the interest of this paper, we consider a random walk where step
length $l$ obeys the distribution:
\begin{equation}
P(l)=\frac{\beta}{l_{0}\left(1+\frac{l}{l_{0}}\right)^{1+\beta}}\label{eq:random-walk}
\end{equation}
This is a normalized version of \cite{gutowski2001evy} with the scale
factor $l_{0}$ added since it is more natural to think in term of
physical dimension of given space. This form preserves the property
of distribution in Eq.~(\ref{eq:Levy}) for large $l$ but it is
much simpler to deal with small $l$. The distribution is heavy-tailed,
as shown on Fig.~\ref{fig:Flight-length-distribution}(a). 

It is not difficult to verify that $l$ can be randomly generated
as follows:
\begin{equation}
l=l_{0}\left(\frac{1}{U^{1/\beta}}-1\right)\label{eq:generate-Levy}
\end{equation}
where $U$ is uniformly distributed in the interval $[0,1)$.

Figs.~\ref{fig:Flight-length-distribution}(b-d) show L\'evy fights
on a 2D landscape with $l_{0}=1$. Note that the scale of Fig.~\ref{fig:Flight-length-distribution}(b)
is much larger than that of Fig.~\ref{fig:Flight-length-distribution}(d)
(around order of 102). For small $\beta$, the random walker tends
to get crowded around a central location and occasionally jumps a
very big step to a new location. As $\beta$ increases, the probability
of performing a long jump decreases. Note that Fig.~\ref{fig:Flight-length-distribution}(d)
shows the random walks as $\beta=3$, which gets outside of the range
given in Eq.~(\ref{eq:Levy}). However, for the computational purposes,
we need not to strictly be consistent with physics laws.

\section{Algorithms}

\subsection{Search mechanisms}

The goal in optimization is to efficiently explore the search space
in order to find globally optimal. L\'evy flights provide a mechanism
to achieve this goal. We use the term ``particle'' to call the abstract
entity that moves on the search landscape. From each position, the
particle can ``fly'' to a new feasible position at a distance, which
is randomly generated from L\'evy flights distribution. It is known
that a good search algorithm often maintains balance between \emph{local
exploitation} and \emph{global exploration} \cite{blum2003metaheuristics}.
L\'evy flights offer a nice way: As in Eq.~(\ref{eq:random-walk}),
we control the frequency and the length of long jumps by adjusting
the parameters $\beta$ and $l_{0}$, respectively. Ideally, algorithms
will dynamically tune these two parameters to best fit a given landscape.

There are several ways to formulate an algorithm that employs such
L\'evy-based distance. Firstly, L\'evy flights define a manageable
move strategy, which can include small local steps, global jumps and
(or) mixing between the two. As the L\'evy process can occasionally
generate long jumps, it can be employed in the multistart model. Secondly,
one can use a single particle (as in Greedy, SA, Tabu Search) \cite{glover1999tabu,talbi2002taxonomy}
or a set of particle(s) (as in GA, Evolution Strategy, Genetic Programming,
ACO and Scatter Search) \cite{blum2003metaheuristics} moving over
the search space. The third way, and probably the most promising,
is to combine generic movement model provided by L\'evy flights with
other known single-solution, or population-based meta-heuristics.
Such combinations could result in more powerful hybrid algorithms
than the originals \cite{talbi2002taxonomy}.

\subsection{LFO algorithms}

Here we propose five algorithms under the class of L\'evy Flights
Optimization (LFO). We first provide the brief description. Pseudocode
for the first four algorithms is then given in Algorithms~\ref{alg:LFO-B}--\ref{alg:LFO-ILS}.

\paragraph{Basic LFO (LFO-B).}

This basic algorithm uses a set of particles at each generation. Starting
from one best-known location, the algorithm will generate a new generation
at distances which are randomly distributed according to L\'evy flights.
The new generation will then be evaluated to select the most promising
one. The process is repeated until stopping criteria are satisfied.
The algorithm is quite closed to GA in a way that it iterates through
generations of particle population, and at each generation it selects
the good individuals for the next step. However, there selection policy
in LFO-B is quite simple: only the best of population survives at
each iteration.

\paragraph{Hybrid LFO with Local Search (LFO-LS).}

This is an extension of LFO-B algorithm, where before the selection
is made each particle performs its own search to reach local optima.
The selection procedure then locates the best local optima found so
far, and L\'evy flights mechanism helps escape from such trap. It
is open to implement the local search algorithm.

\paragraph{Local Search with Multiple LFO Restarts (LFO-MLS).}

This can be viewed as a sequential version of LFO-LS algorithm and
actually behaves as a Multi-start Local Search. Here only one particle
is used. In each iteration the particle tries local search until getting
trapped in local optima. To escape from such trap, it will jump to
new location by a step generated from L\'evy distribution. The jump
is immediately accepted without further selection.

\paragraph{LFO with Iterated Local Search (LFO-ILS).}

This is very similar to LFO-MLS except for the local minima escape
strategy. Instead of immediately accepting the L\'evy-based jump,
the algorithm tries a number of jumps until a better solution is found.
This behavior is identified as Iterated Local Search \cite{Lourenco-et-al03}.

\paragraph{LFO-MLS + SA (LFO-SA).}

This is a combination of LFO-MLS and SA separated in time: LFO-MLS
is run first, then the SA takes the solution as the starting point.
The idea is that LFO algorithms may locate the good solution quickly
while SA helps improve the quality in the long run. 

\begin{algorithm}
\textbf{procedure:} LFO\_B()

$\quad$\emph{init\_position}();

$\quad$\textbf{while} (stopping\_criteria\_not\_met)

$\quad$$\quad$\textbf{for} each member in the new generation

$\quad$$\quad$$\quad$$l$\emph{$\leftarrow$L}\'e\emph{vy\_flights}($l_{0}$,$\beta$); 

$\quad$$\quad$$\quad$\emph{jump\_randomly\_at\_distance}($l$); 

$\quad$$\quad$\textbf{endfor}

$\quad$$\quad$\emph{return\_to\_best\_known\_position}();

$\quad$\textbf{endwhile}

\textbf{endprocedure}

\caption{Basic LFO (LFO-B). \label{alg:LFO-B}}
\end{algorithm}

\begin{algorithm}
\textbf{procedure:} LFO\_LS()

$\quad$\emph{init\_position}();

$\quad$\textbf{while} (stopping\_criteria\_not\_met)

$\quad$$\quad$\textbf{for} each member in the new generation

$\quad$$\quad$$\quad$$l$\emph{$\leftarrow$L}\'e\emph{vy\_flights}($l_{0}$,$\beta$); 

$\quad$$\quad$$\quad$\emph{jump\_randomly\_at\_distance}($l$);

$\quad$$\quad$$\quad$\emph{perform\_local\_search}();

$\quad$$\quad$\textbf{endfor}

$\quad$$\quad$\emph{return\_to\_best\_known\_position}();

$\quad$\textbf{endwhile}

\textbf{endprocedure}

\caption{Hybrid LFO with Local Search (LFO-LS). \label{alg:LFO-LS}}
\end{algorithm}

\begin{algorithm}
\textbf{procedure:} LFO\_MLS()

$\quad$\emph{init\_position}();

$\quad$\textbf{while} (stopping\_criteria\_not\_met)

$\quad$$\quad$\emph{perform\_local\_search}();

$\quad$$\quad$\emph{$l$$\leftarrow$L}\'e\emph{vy\_flights}($l_{0}$,$\beta$);

$\quad$$\quad$\emph{jump\_randomly\_at\_distance}($l$);

$\quad$\textbf{endwhile}

\textbf{endprocedure}

\caption{Local Search with Multiple LFO Restarts (LFO-MLS). \label{alg:LFO-MLS}}
\end{algorithm}

\begin{algorithm}
\textbf{procedure:} LFO\_ILS()

$\quad$\emph{init\_position}();

$\quad$\textbf{while} (stopping\_criteria\_not\_met)

$\quad$$\quad$\emph{perform\_local\_search}();

$\quad$$\quad$\emph{$l$$\leftarrow$L}\'e\emph{vy\_flights}($l_{0}$,$\beta$);

$\quad$$\quad$\textbf{while}(not found better solution)

$\quad$$\quad$$\quad$\emph{jump\_randomly\_at\_distance}($l$);

$\quad$$\quad$\textbf{endwhile}

$\quad$\textbf{endwhile}

\textbf{endprocedure}

\caption{LFO with Iterated Local Search (LFO-ILS). \label{alg:LFO-ILS}}
\end{algorithm}

\section{Experiments and Results}

\subsection{Test problems}

Test problems are $f_{0}$ of \cite{corana1987minimizing}, $f_{2}$
(Rosenbrock\textquoteright{}s saddle) and $f_{5}$ (Shekel\textquoteright{}s
foxholes) in De Jong\textquoteright{}s test function suite \cite{De-Jong75},
Rastrigin\textquoteright{}s $f_{6}$ \cite{whitley1995building} and
Keane\textquoteright{}s Bump \cite{keane1996brief}.

\paragraph{$f_{0}$:}

\begin{eqnarray*}
f_{0}(x_{1},x_{2},x_{3},x_{4}) & = & \sum_{i=1}^{4}\begin{cases}
\left(t_{i}\mbox{sgn}(z_{i})+z_{i}\right)^{2}cd_{i} & \,\,\mbox{if}\,\left|x_{i}-z_{i}\right|<\left|t_{i}\right|\\
d_{i}x_{i}^{2} & \,\,\mbox{otherwise}
\end{cases}\\
\mbox{where:}\\
z_{i} & = & \left\lfloor \left|\frac{x_{i}}{s_{i}}\right|+0.49999\right\rfloor \mbox{sgn}(x_{i})s_{i}\\
s_{i} & = & 0.2,\, t_{i}=0.05\\
d_{i} & = & \left\{ 1,1000,10,100\right\} \\
c & = & 0.15\\
-1000 & \le & x_{i}\le1000.
\end{eqnarray*}

\paragraph{$f_{2}$:}

\begin{eqnarray*}
f_{2}(x_{1},x_{2},...,x_{N}) & = & \sum_{i=1}^{N-1}\left(100\left(x_{i}^{2}-x_{i+1}\right)^{2}+\left(1-x_{i}\right)^{2}\right)\\
\mbox{where:}\\
-2.048 & \le & x_{i}\le2.048,\,\, i=1,2,...,N.
\end{eqnarray*}

\paragraph{$f_{5}$:}

\begin{eqnarray*}
f_{5}(x_{1},x_{2}) & = & \frac{1}{\frac{1}{500}+\sum_{j=1}^{25}\frac{1}{j+\sum_{i=1}^{2}(x_{i}-a_{ij})^{6}}}\\
a_{1j} & = & \{-32,-16,0,16,32,-32,-16,0,16,32,-32,-16,0,16,32,\\
 &  & \quad-32,-16,0,16,32,-32,-16,0,16,32\}\\
a_{2j} & = & \{-32,-32,-32,-32,-32,-16,-16,-16,-16,-16,0,0,0,0,0,\\
 &  & \quad16,16,16,16,16,32,32,32,32,32\}\\
-65.536 & \le & x_{i}\le65.536,\, i=1,2.
\end{eqnarray*}

\paragraph{$f_{6}$:}

\begin{eqnarray*}
f_{6}(x_{1},x_{2},...,x_{N}) & = & 10N+\sum_{i=1}^{N}\left(x_{i}^{2}-10\cos(2\pi x_{i})\right)\\
-5.12 & \le & x_{i}\le5.12
\end{eqnarray*}

\paragraph{Bump:}

\begin{eqnarray*}
f_{BUMP}(x_{1},x_{2},...,x_{N}) & = & 1-\frac{\left|\sum_{i=1}^{N}\cos^{4}(x_{i})-2\prod_{i=1}^{N}\cos^{2}(x_{i})\right|}{\sqrt{\sum_{i=1}^{N}ix_{i}^{2}}}\\
0 & < & x_{i}<10;\,\prod_{i=1}^{N}x_{i}>0.75;\,\sum_{i=1}^{N}x_{i}<\frac{15N}{2}
\end{eqnarray*}

Function $f_{0}$ is very difficult to minimize with $10^{20}$ local
minima, and $f_{2}$ is also considered to be hard despite of being
uni-modal \cite{corana1987minimizing}. $f_{6}$ is interesting because
of the sinuous component. According to Keane \cite{keane1996brief},
the Bump is a seriously hard test-function for optimization algorithms
because the landscape surface is highly ``bumpy'' with very similar
peaks and global optimum is generally determined by the product constraint.
Among those functions, $f_{0}$ and $f_{5}$ have fixed dimensions
while the rest can be set freely. In our tests, the dimension of 10
is set for $f_{2}$ and $f_{6}$ and of 50 for the Bump.

\subsection{Algorithms implementation}

We compare four proposed algorithm with the classic SA. This subsection
provides greater details in algorithm realization, which can be classified
into \emph{move strategy}, \emph{stopping criteria}, and \emph{algorithm
specificity}.

\paragraph{Move strategy.}

Each move is selected in random directions spanning in all dimensions
of search space where move length is measured in Euclidean metric.
In the case of infeasible move to the region outside bounded space,
two strategies are used: (i) the move selection is repeated until
a feasible one is found, and (ii) the move is stopped at the edges.
The first strategy is quite intuitive but it may result in many repetitions
in case of long L\'evy flights. The second gives chance to explore
the boundary regions, in which high quality solutions may be found.

\paragraph{Stopping criteria.}

The main stopping criteria used in main algorithms are time and the
quality of best solution found so far (compared with known optimal
one). Another stopping criterion is number of non-improvement moves.
This can be either number of steps in neighborhood exploitation or
number of jumps in global exploration. The first is used in greedy
search, while the second applies for multiple restart type.

\paragraph{Algorithm specificity.}

In SA, the initial temperature $T_{0}$ is chosen as 10\% of randomly
generated initial solution while the stopping temperature $T_{s}$
is fixed at $0.0001$. Although there are no exact reasons for such
choice, it is based on author\textquoteright{}s experience with SA
so that transition probability is always 1 at the beginning and very
close to 0 at the end of each run. The cooling schedule is the widely
used power type:
\[
T(t)=r^{t}T_{0}\,\,\mbox{where}\,\, r=e^{\mbox{ln}(T/T_{0})/t_{m}}
\]
and $t_{m}$ is the maximum allocated run time. In all LFO algorithms,
we use the power index $\beta=1.5$. The number of jumps is set at
100 for LFO-B and LFO-MLS algorithms, while the jump distance is limited
at a half of largest size of search space\textquoteright{}s dimensions.

\subsection{Results and Discussions}

\begin{figure}
\begin{centering}
\includegraphics[width=0.7\textwidth]{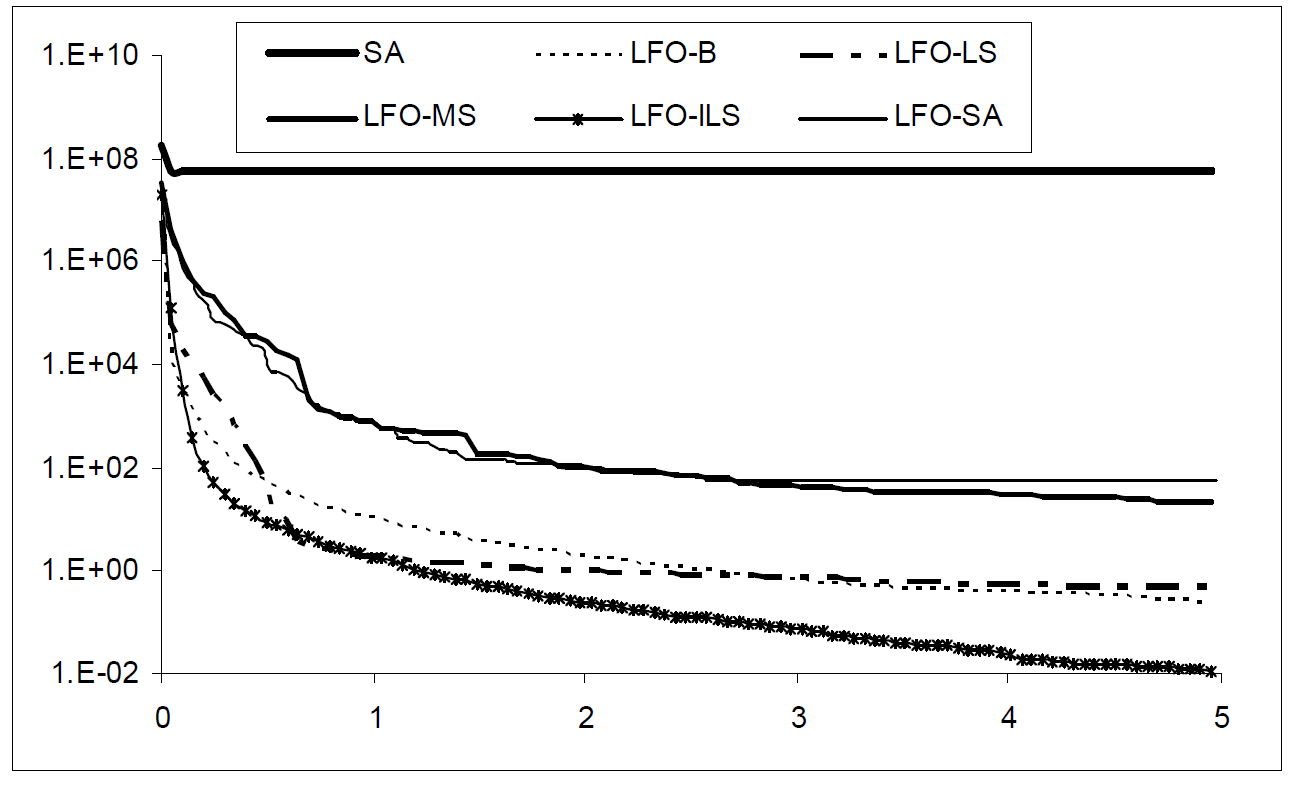}
\par\end{centering}

\caption{Test results for $f_{0}$.\label{fig:Test-results-for-F0}}

\end{figure}

\begin{figure}
\begin{centering}
\includegraphics[width=0.7\textwidth]{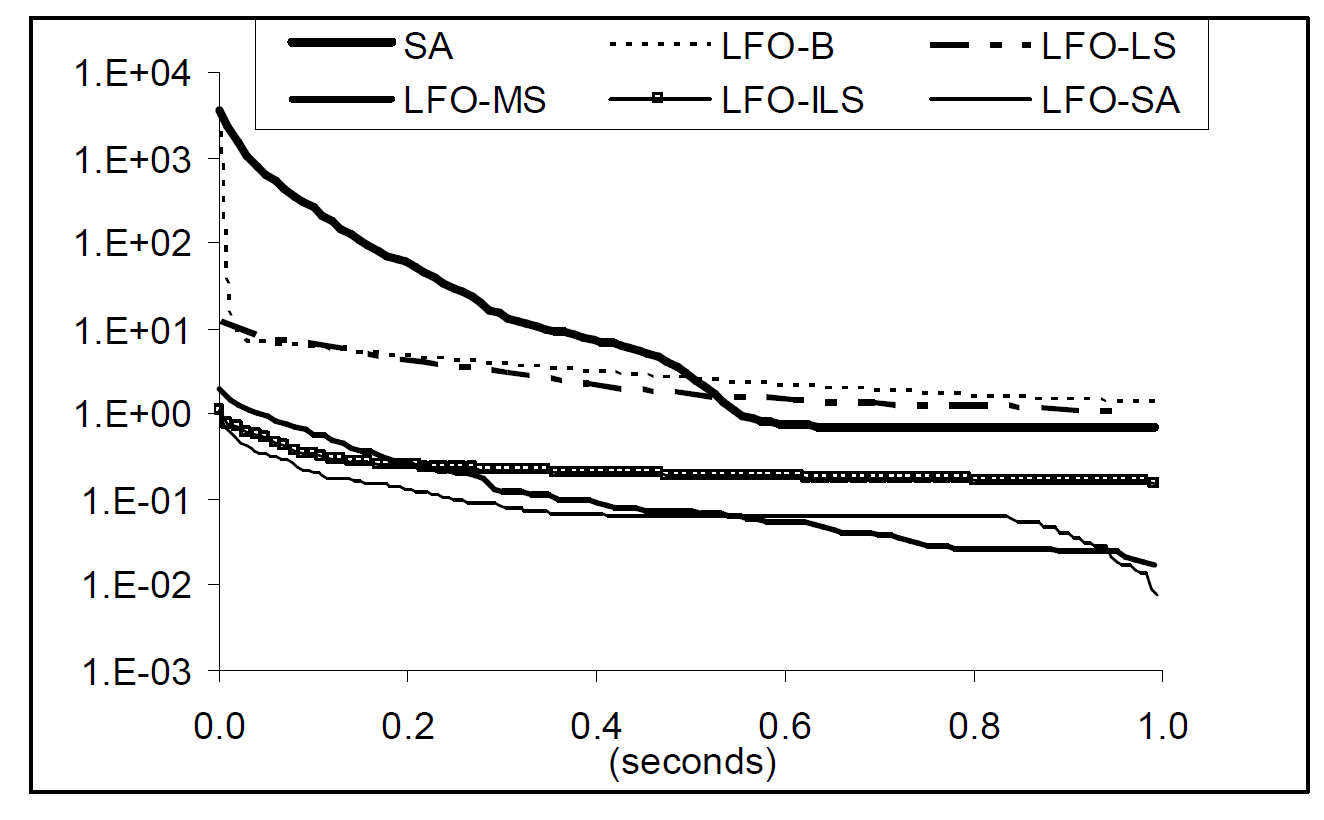}
\par\end{centering}

\caption{Test results for $f_{2}$.\label{fig:Test-results-for-F2}}
\end{figure}

\begin{figure}
\begin{centering}
\includegraphics[width=0.7\textwidth]{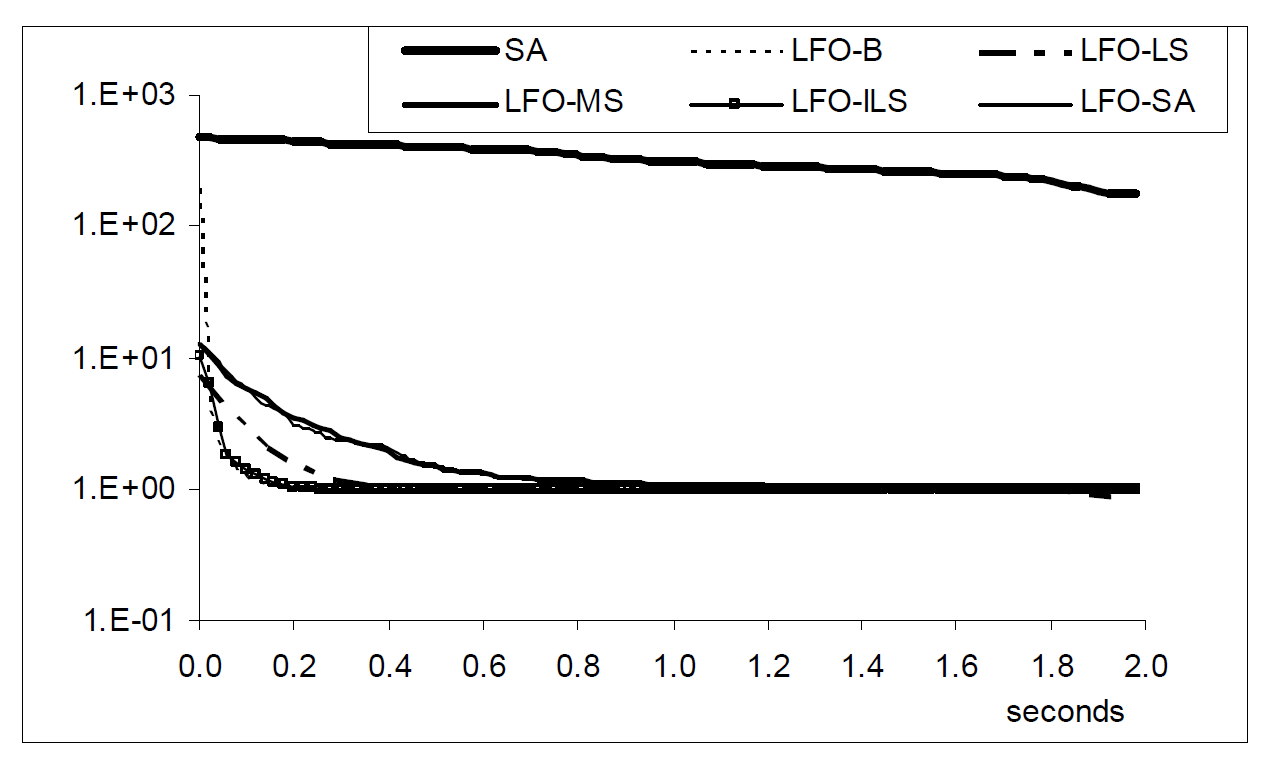}
\par\end{centering}

\caption{Test results for $f_{5}$.\label{fig:Test-results-for-F5}}
\end{figure}

\begin{figure}
\begin{centering}
\includegraphics[width=0.7\textwidth]{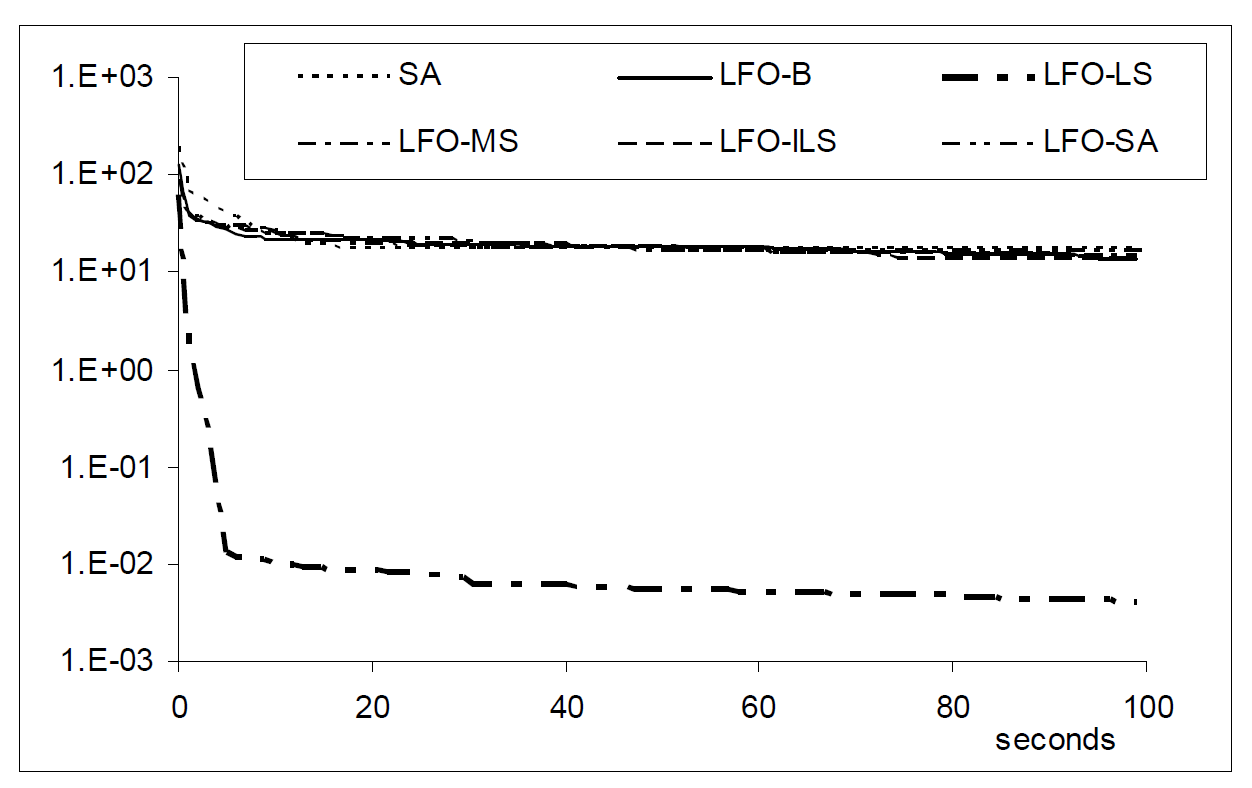}
\par\end{centering}

\caption{Test results for $f_{6}$.\label{fig:Test-results-for-F6}}
\end{figure}

\begin{figure}
\begin{centering}
\includegraphics[width=0.7\textwidth]{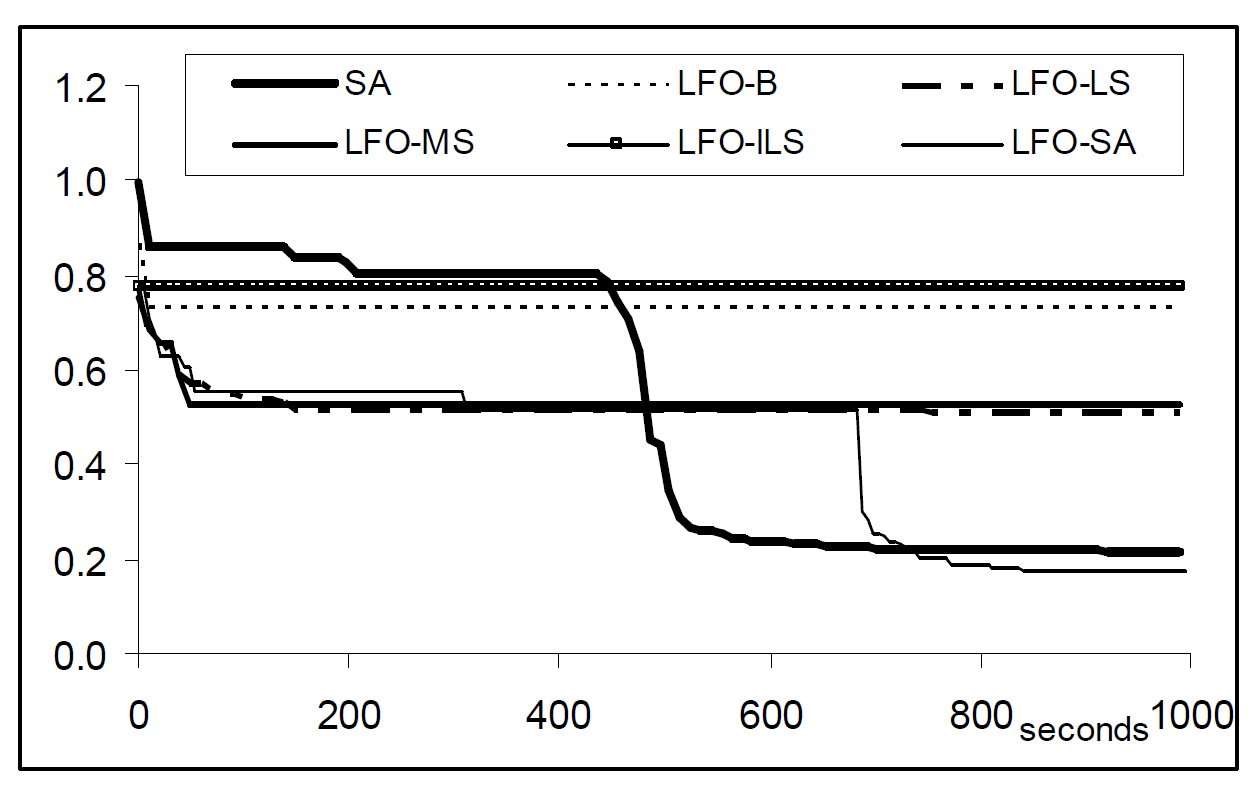}
\par\end{centering}

\caption{Test results for $f_{6}$.\label{fig:Test-results-for-Bump}}
\end{figure}

The tests were run on 2.5GHz Intel computer within given periods of
time. Every test was run repeatedly and best found solutions were
sampled at relevant intervals and then averaged. Numbers of replications
were 100 times for $f_{0}$, $f_{2}$, and $f_{5}$, 20 times for
$f_{6}$ and 10 times for the Bump. Algorithms\textquoteright{} performances
on the five test problems are presented in Figs.~\ref{fig:Test-results-for-F0}--\ref{fig:Test-results-for-Bump}.

In all cases, some of LFO algorithms outperform SA although SA appears
to improve its performance in the long runs. This is a proven beauty
of SA but it will be impractical if the required time is too long
for daily activities. LFO algorithms tend to sample good quality solutions
(compared to initial random solution) very quickly, in most cases
within a second. The only exception is the Bump problem, where SA
seems to work best after a significant time.

Although there are not enough representative test cases to statistically
conclude on the power of LFO algorithms over SA, there are several
interesting points to note. Firstly, L\'evy flights portrays well
the distribution of flight lengths and times performed by foragers
observed in natural experiments \cite{viswanathan1999optimizing}.
In the process of foraging over a given landscape, L\'evy distribution
helps reduce the probability of returning previously visited locations,
compared to normal distribution. This is particularly advantageous
in memoryless algorithms like basic stochastic SA or GA, where there
are no built-in mechanisms to avoid revisits.

Secondly, as mentioned early in this paper, LFO algorithms work under
assumption that good quality solutions can be found around local minima.
Such property can be found in most test cases, where experiments have
proved the algorithms\textquoteright{} favor. However, in the Bump
problem, such ``big valley'' hypothesis does not hold because the
global minimum is located on search space boundary. This property
can help explain why LFO algorithms fail to model the Bump landscape
structure but succeed in other cases.

Finally, the hybrid LFO-SA appears to work as well as or better SA
in all cases while keeping the similar search power to other LFO algorithms.
It is obvious that LFO-SA behaves exactly like LFO-MLS in the short
run (in $f_{0}$, $f_{2}$, $f_{5}$ and $f_{6}$, see Figs.~\ref{fig:Test-results-for-F0}--\ref{fig:Test-results-for-F6})
and like SA in the long course (in Bump, see Fig.~\ref{fig:Test-results-for-Bump}).

\section{Conclusions \& outlooks}

The paper has proposed a set of optimization algorithms based on L\'evy
flights. The algorithms were validated against Simulated Annealing
on several hard continuous test functions. Experiments have demonstrated
that LFO (L\'evy Flights Optimization) could be superior to SA. 

Algorithms implemented in this paper are rather basic, and there are
rooms for further extension. For example, the two main parameters
of mean L\'evy distance and the power index can be adjusted dynamically
during the run. Or one may wish to extend the idea of Particle Swarm
Optimization (PSO) in the way that the whole population keeps continuous
flying while dynamically adjusting speed and direction based on information
collected so far. Even each flying particle can be treated as an autonomous
agent involving collective intelligent decision making. For those
who are familiar with GA, it is possible to extend the practice of
GA in the way that at each generation, we select a set of particles
in stead of the best one to form the next generation.


\end{document}